\documentclass{article}

\AtBeginDocument{%
  \providecommand\BibTeX{{%
    \normalfont B\kern-0.5em{\scshape i\kern-0.25em b}\kern-0.8em\TeX}}}

\usepackage[a4paper, total={6in, 8in}]{geometry}
\usepackage{comment}

\usepackage[utf8]{inputenc}
\usepackage[english]{babel}
\usepackage{natbib}
\setcitestyle{authoryear,open={(},close={)}}
\usepackage[size=small,format=plain,labelfont=up,textfont=up]{caption}
\usepackage{array, booktabs}
\usepackage{color}

\definecolor{darkblue}{rgb}{0.0,0.0,0.5}
\definecolor{lightblue}{rgb}{0.0,0.0,0.8}
\definecolor{LightSteelBlue3}{rgb}{0.635,0.71, 0.804}

\usepackage{hyperref}

\hypersetup{
	colorlinks,breaklinks,linkcolor=lightblue,
    urlcolor=lightblue,
	anchorcolor=lightblue,
    citecolor=darkblue
} 

\newcommand{\foo}{\color{black}\makebox[0pt]{\textbullet}\hskip-0.5pt\vrule width 1pt\hspace{\labelsep}}

\newcommand{\II}{{\mathcal I}}

\usepackage{amsmath,amssymb,amsthm,bm}
\usepackage{thmtools,thm-restate}

\newcommand{\V}[1]{{\mathbf{#1}}}

\newcommand{\abs}[1]{ {\left| #1 \right|}}

\title{Interpretable Artificial Intelligence through the Lens of Feature Interaction}
\author{Michael Tsang, James Enouen, Yan Liu\\University of Southern California}

\date{}

\usepackage{graphicx}

\begin{document}

\maketitle

\section{Introduction}
Deep learning, alongside other modern machine learning techniques, has become the state of the art solution for a diverse range of real-world tasks.
These include a variety of sensitive applications such as healthcare, finance, autonomous driving, criminal justice, and others which all pose significant concerns for fairness, robustness, safety, and trustworthiness.
Despite these applications to critical tasks, deep networks are infamously referred to as black-box models because of their total lack of transparency in decision-making.
If we are able to gain insight into how a model is coming to its conclusions, we are able to more clearly assess the trustworthiness and validity of its decisions.
Consequently, an abundance of ongoing research is attempting to address model interpretability as the key problem to resolving these issues.



There are many methods which are currently used to provide explanations of complex model predictions.
LIME \citep{ribeiro2017lime} fits a local linear model around a data point, showing which features positively and negatively influence the prediction results.
Despite the overall model being nonlinear, the local model gets an interpretable picture of how the model looks at small scales around the data point.
Extensions of this method use other interpretable models like small decision tress.
Shapley Values and SHAP follows a similar idea to assign a score to each feature, using a game-theoretic formulation which treats each feature as a player causing the final prediction \citep{lundberg2017unified}.
Its more rigorous formulation yields guarantees of its explanations summing up to the prediction score, but practically it usually must be estimated because of its high computational cost.
Shuffle-based feature importance permutes the data of each feature to ascertain its importance in the final prediction in comparison to its normal prediction \citep{fisher2018model}.
IG uses the fundamental theorem of calculus to provide additive explanations of a prediction \citep{sundararajan2017axiomatic}.
This method is very popular in computer vision where its computational efficiency and saliency are prized, even though other work has exposed some of its shortcomings in providing an interpretation \citep{adebayo2018saliencysanity}.
Other methods are specifically designed for computer vision like TCAV \citep{kim2018interpretability} which finds a `concept direction' corresponding to a large sample of concept images from the user.
Surprisingly, all of these most popular interpretability methods share the same one limitation.
None of these methods consider the shared importance of groups of two or more features; they only look at the effects had by each of the features individually.

A {feature interaction} between two variables broadly describes a situation where both of the features/ variables are \textit{simultaneously} important for a model's prediction.
In text applications for sentiment, "not good" is a very simple example of two words strongly interacting with one another to create a negative sentiment.
In modern-day applications, neural networks are usually hailed as amazing function approximators exactly because of their incredible ability to {automatically} uncover these kinds of complex relationships between the variables of the dataset.
In many ways, however, this automatic discovery is what leads to deep networks having totally unintelligible decision-making processes.
Modern research has made it increasingly clear that state-of-the-art models do in fact depend on these feature interactions for their good performance, so any full explanation of these models' decisions must include the interactions between features.
This worryingly concludes that many of our interpretable explanations fail to accurately represent the model and we must choose between interpretability and complexity.
Accordingly, the perspective of feature interactions provides the necessary framework to study both discovering and explaining these complex interactions, offering the potential to build a bridge between the simple, interpretable model and the powerful, complex one.

This survey aims to capture the progress on interpreting feature interactions given the recent advantages of deep learning and complex prediction models.
After we first provide a recap of the terminology and initial research, we then survey more recent works on interpreting feature interactions in modern prediction models like neural networks and random forests.
Specifically, this survey is outlined as follows: 

We begin with a discussion on what is a feature interaction in~\S\ref{sec:interaction} and what is interpretability in~\S\ref{sec:interpretability}. 
In~\S\ref{sec:history}, we survey historical feature interaction detection methods.
In~\S\ref{sec:moderndiscussion}, we overview modern techniques from the two primary lenses on feature interactions: ``detection'' and ``interpretation.''
We then survey the many applications of feature interaction interpretation in~\S\ref{sec:application}. 
Lastly, we offer thoughts on future research in~\S\ref{sec:future}.

Under the purview of feature interactions, explaining a full model is able to boil down to explaining the model's most salient feature interactions.

In addition, existing work at the intersection of these two fields has not only shown that deep models can assist feature interaction detection of the underlying data distribution, but that feature interactions can thereafter be used to push model performance beyond that of the original neural network.

Consequently, it has become clear that further studying this intersection has the potential to grant a better understanding of deep learning overall.

\section{What is a Feature Interaction?}
\label{sec:interaction}

A feature interaction 
describes a situation in which the effect of one feature on an outcome depends on the state of a second feature.
For example, consider a linear regression model 
$f(\V{x})= w_1x_1 + w_2x_2 + w_{3}x_1x_2$, where $x_1$ and $x_2$ are features and $\{w_i\}$ are coefficients. The multiplication $x_1x_2$ forms a feature interaction, and the individual terms $x_1$ and $x_2$ are main effects. The coefficients provide information about each term's importance. A feature interaction is not restricted to a multiplicative effect between features, nor does it have to be between only two features. In general, a feature interaction is any non-additive effect between multiple features on an outcome.

The most common formal definition of feature interaction is as follows:
\begin{restatable}[Statistical Non-Additive Interaction]{definition}{interaction}\label{def:interaction}
A function $f$ contains a statistical non-additive interaction of multiple features indexed in set $\II$
if and only if $f$ cannot be decomposed into a sum of $|\II|$ subfunctions $f_i$ , each excluding the $i$-th interaction variable:
    $f(\V{x}) \neq \sum_{i\in\II} f_i(\V{x}_{\setminus \{i\}})$.
\end{restatable}
Def.~\ref{def:interaction} identifies a non-additive effect among all features $\mathcal{I}$ on the output of function $f$~\citep{friedman2008predictive, sorokina2008detecting, tsang2018detecting}.
For example, this means that 
the function $\sin(x_1+x_2)$
creates a feature interaction
because it cannot be represented as an 
addition of univariate functions, i.e., 
 $\sin(x_1+x_2)\neq f_1(x_2)$ + $f_2(x_1)$. 
 We refer to individual feature effects which do not interact with other features as \emph{main effects}. Higher-order feature interactions are captured by $\abs{\II}>2$, i.e. interactions larger than pairs. Additionally, if a higher-order interaction exists,  all of its subsets also exist as interactions~\citep{sorokina2008detecting, tsang2018detecting}.


Feature interaction phenomena exist in many real-world settings where an outcome is modeled as a function of features. Because machine learning is precisely designed to model such functions, high performance prediction  models - e.g. neural networks - are well suited for learning feature interactions. Feature interactions manifest beyond tabular data settings too, such as word interactions in text-based sentiment analysis and image-patch interactions in image classification. Examples of word interactions are ``not, good'' and ``not, bad'' in movie reviews. An example of an image interaction is a combination of image patches corresponding to the ears, nose, and eyes of a dog, which supports a dog classification. We see that interaction phenomena can be interpretable in diverse domains.

\section{What is Interpretability?}
\label{sec:interpretability}

 Model interpretability is the ``degree to which an observer can understand the causes of a prediction''~\citep{biran2017explanation,miller2019explanation}.
There are many different factors determine how effective a given interpretation is at providing an explanation.
\begin{itemize}
    \item \textbf{The observer}: The person observing the explanation may belong to many different audiences.  A domain expert will always have more intuition on a subject than an average person.  We may or we may not want to assume the observer has background knowledge.
    \item \textbf{The goal}: The observer likely has an objective, explicit or implicit, when they look for an explanation from a model.  For example, a data scientist might be looking for shortcomings in the model; a statistician could be looking for patterns in the data; a doctor could be hoping for sanity in the model's predictions; etc.
    \item \textbf{The complexity}: The model's underlying complexity and consequently the ``true" explanation's complexity can become intractable to compute.  Moreover, explaining exactly why a model made its decision may not be preferable to finding explanations that align with our intuition.  It is possible there won't always be a middle ground.
    \item \textbf{The structure}: How the explanation is organized and the actual content it provides has a great effect on how it is ultimately perceived.  Some examples are listing important features, plotting input-output trends, providing a distilled, simpler explanation model, etc.  It is very likely that the explanation's structure will vary across applications and alongside these other factors.
\end{itemize}

For a relevant discussion on the social science behind human understanding, we refer readers to the excellent survey by~\citet{miller2019explanation}. 
In this survey, they discuss how the interpretation of feature interactions can be informative and comprehensible to audiences with little background knowledge. 
Common questions investigated by interaction interpretation methods  are \emph{what} features and \emph{how} feature interactions affect predictions. 
A great resource for more methods of interpretability can be found in \citet{molnar2019}.

\section{A History of Feature Interactions}
\label{sec:history}

A common thread among  historical research on feature interactions is that most methods  were developed to \emph{detect} feature interactions in the data.
Feature interaction detection addresses the question of \emph{which} feature interactions (out of all possible subsets) are the ones that actually affect outcomes.
Often, this process leverages a trained model's interpretations of features as a surrogate for interaction detection.

\subsection{Timeline} The notion of a feature interaction has been studied at least since the $19$th century when John Lawes and Joseph Gilbert used factorial designs in agricultural research at the Rothamsted Experimental Station~\citep{dean2015handbook}. A factorial design is an experiment that includes observations at all combinations of categories of each factor or feature.
However, the ``advantages [of factorial design] had never been clearly
recognised, and many research workers believed that the best course was
the conceptually simple one of investigating one question at a time''~\citep{yates1964sir}. In the early $20$th century, \citet{fisher1926048} emphasized the importance of factorial designs as being the only way to obtain information about feature interactions. Near the same time,~\citet{fisher1921probable} also developed one of the foundations of statistical analysis called Analysis of Variance (ANOVA) including two-way ANOVA~\citep{fisher1925statistical}, which is a factorial method to detect pairwise feature interactions based  on differences among group means in a dataset. \citet{tukey1949one} extended two-way ANOVA to test if two categorical features are non-additively related to the expected value of a outcome variable. This work set a precedent for later research on detecting feature interactions based on their non-additive definition. Soon after,  experimental designs were generalized  to study feature interactions, in particular the generalized randomized block design~\citep{wilk1955randomization}, which assigns test subjects to different categories (or blocks) between features in a way where cross-categories between features serve as interaction terms in  linear regression.

There was a surge of interest in improving the analysis of feature interactions after the mid $20$th century.  \citet{belson1959matching,morgan1963problems} proposed Automatic Interaction Detection (AID) originally under a different name. AID detects interactions by subdividing data into disjoint exhaustive subsets to model an outcome  based on categorical features. Based on AID, \citet{kass1980exploratory} developed Chi-square Automatic Interaction Detection (CHAID), which determines how categorical features best combine in decision trees via a chi-square test. AID and CHAID were precursors to modern decision tree prediction models. Concurrently, \citet{nelder1977reformulation} introduced the ``Principle of Marginality'' arguing that a feature interaction and its marginal variables should not be considered separately, for example in linear regression. \citet{hamada1992analysis} provided a contrasting view  that an interaction is only important if one or both of its marginal variables are important. At the start of the $21$st century, efforts began to focus on interpreting interactions in accurate prediction models. \citet{ai2003interaction} proposed extracting interactions from logit and probit models via mixed partial derivatives. \cite{gevrey2006two} followed up by proposing mixed partial derivatives to extract interactions from multilayer perceptrons with sigmoid activations when at the time, only shallow neural networks were studied. \citet{friedman2008predictive} proposed using hybrid models to capture interactions with decision trees and univariate effects with linear regression. \citet{sorokina2008detecting} proposed to use high-performance additive trees to detect feature interactions based on their non-additive definition. At the turn of the decade, we saw~\citet{bien2013lasso} capture interactions with different heredity conditions using a hierarchical lasso on linear regression models. Then, \citet{hao2014interaction} drew attention towards interaction screening in high dimensional data. As deep learning became popular in this decade,~\citet{tsang2018detecting} found that feature interactions can be detected from neural network weights. Feature interaction detection was also applied to general prediction models, e.g. deep neural networks~\citep{tsang2020feature,tsang2020does}. We provide a timeline for this research history in Table~\ref{table:timeline}.

{
\renewcommand\arraystretch{1.6}

\begin{table}
\centering
\captionsetup{singlelinecheck=false, labelfont=sc, labelsep=quad}
\center{
\caption{Timeline of feature interaction history\label{table:timeline}}
}\vskip -1.5ex
\begin{tabular}{p{6cm} @{\hskip 9pt}  @{\,}r <{\hskip 2pt} !{\foo} >{\raggedright\arraybackslash}p{6cm}}
\toprule
\addlinespace[1.5ex]
\emph{Lawes \& Gilbert} - factorial design in agricultural research at the Rothamsted Experimental Station &$1843$ & \\
\emph{Fisher} -  two-way Analysis of Variance (ANOVA) & $1925$ & \\
&$1949$ & \emph{Tukey} - Tukey’s test of additivity\\
&$1955$ & \emph{Wilk} - generalized random block design\\
\emph{Belson} - Automatic Interaction Detection by subdividing data &$1959$ & \\
\emph{Nelder} - Principle of Marginality&$1977$ & \\
&$1980$ & \emph{Kass} - Chi-square Automatic Interaction Detection by combining features in decision trees via chi-square tests\\

&$1991$ & \emph{Aiken \& West} - book on interpreting interaction effects\\
\emph{Hamada \& Wu} - heredity conditions &$1992$ & \\
 \emph{Ai \& Norton} -  interactions in logit and probit models &$2003$ &\\
  &$2006$ & \emph{Gevry et al.} - interactions in sigmoid neural networks\\
\emph{Friedman \& Popescu} - RuleFit to detect interactions by mixing linear regression and trees &$2008$ & \emph{Sorokina et al.} - Additive Groves to detect non-additive interactions \\
 \emph{Bien et al.} - Hierarchical Lasso&$2013$ & \\
\emph{Hao \& Zhang} - interaction screening in high dimensional data &$2014$ & \\
&$2018$ & \emph{Tsang et al.} - interaction detection from neural network weights\\
\end{tabular}
\end{table}
}
\renewcommand\arraystretch{1}

\newpage

\section{Modern Discussion of Feature Interactions}
\label{sec:moderndiscussion}
There are two main objectives throughout the domain of feature interaction.
The classical objective is to \emph{find} the groups of features which depend on one another: ``Feature Interaction Detection.''
The other main objective is, given a specific set of features, to \emph{understand} in what way this group of features interacts with one another: ``Feature Interaction Interpretation."

\subsection{Feature Interaction Detection}

\subsubsection{Individual Testing}
The most common form of interaction detection tests each combination of features separately. This individual testing approach started with ANOVA~\citep{fisher1925statistical} and has continued with modern interaction detectors. ANOVA  conducts hypothesis tests for each interaction candidate by checking each hypothesis with F-statistics~\citep{wonnacott1972introductory}. Multi-way ANOVA exists to detect interactions of higher-orders combinations, not just between pairs of features. 
A significant problem with individual testing methods is that they require an exponentially growing number of tests as the desired interaction order increases.
Not only is this approach intractable, but also has a high chance of generating false positives or a high false discovery rate~\citep{benjamini1995controlling} that arises from multiple testing.

\paragraph{Model-Based Testing:}
One type of individual testing method is to train a model both with and without access to an interaction and calculate the difference in performance.
This was exactly the approach of the Additive Groves (AG) method~\citep{sorokina2008detecting}.
This method is notable because it detects interactions using high-performance random forest models while accounting for the  non-additive definition of feature interaction.
This essentially allows the detected feature interaction to be unrestricted to a functional form (e.g. multiplication $x_i x_j$) which was a big step forward at the time. 
AG tests each of these interactions by comparing two regression trees, one that fits all interactions, and the other that has the interaction of interest forcibly removed. 
The main issue with this method is it exacerbates the slowness of individual testing methods which require many tests by forcing a complex model to be trained for each test.

\paragraph{Gradient-Based Testing:}
A prominent approach to detect feature interactions is based on mixed partial derivatives~\citep{friedman2008predictive}, which also uses individual testing.
Namely, the mixed derivative definition of feature interactions is the following:
a function $f(\cdot)$ exhibits statistical interaction $\II$ among all features $x_i$ indexed by
${i_1,i_2,\dots,i_{\abs{\mathcal{I}}}}\in \II$ if
\begin{align}
\mathbb{E}_\V{x}\left[\frac{\partial^{\abs{\mathcal{I}}} f(\V{x})}{\partial x_{i_1}\partial x_{i_2} \dots\partial x_{i_{\abs{\mathcal{I}}}}}\right]^2 > 0.
 \label{eq:grad_interaction}
\end{align}
This definition has been popular to use with logit~\citep{ai2003interaction}, nonlinear~\citep{karaca2012interaction}, and traditional feedforward neural network models~\citep{gevrey2006two}. The advantage of these mixed derivative approaches is that they are exact at interaction detection up to the performance of the model. For example, neural networks excel at interaction detection because they are high performance models as universal function approximators. However,~\eqref{eq:grad_interaction} computes an expectation that can be computationally expensive by requiring many individual tests for just one tested interaction. A theoretical problem with~\eqref{eq:grad_interaction} is that it does not apply to all models, such as
 neural networks with piecewise-linear activation functions including the common ReLU function~\citep{glorot2011deep}.

\subsubsection{Lasso Selection} An alternative approach to interaction detection is based on lasso selection, which is fast and may not require individual testing. One can construct an additive model with many different interaction terms and let lasso shrink the coefficients of unimportant terms to zero~\citep{tibshirani1996regression,bien2013lasso,min2014interpretable,purushotham2014factorized}. While lasso methods are fast, they require specifying all interaction terms of interest. For pairwise interaction detection, this requires $\mathcal{O}(p^2)$ terms (where $p$ is the number of features), and $\mathcal{O}(2^p)$ terms for higher-order interaction detection.
This can be challenging because the lasso-based methods will only discover interactions of the pre-specified forms.

\subsubsection{Neural Interaction Detection}
The modern work of Neural Interaction Detection (NID)~\citet{tsang2018detecting} circumvented this problem by using the powerful inductions of feedforward neural networks.
This work takes advantage of the specific structure of neural networks to avoid needing to specifying large numbers of interactions while still not focusing on individual testing.
The method traces high-strength weights from features to common hidden units as a way to automatically detect which features have high dependencies on one another.
Later work ~\cite{tsang2020feature} uses NID to train a new network with additional `cross-features' from the detected interactions to actually boost the deep networks performance.
Given the popularity of deep learning, the prospect of improving deep networks' performance through understanding feature interactions warrants further investigation into this re-emerging topic of feature interactions.

\subsubsection{Bayesian Group Expected Hessian}
This modern work~\citep{cui2020bayesGEH} also tries to circumvent the issue of detecting general interactions without multiple testing or many specifications.
To do this, they train a Bayesian Neural Network (BNN) with modern variational inference techniques.
They then apply their new detection technique called Bayesian Group Expected Hessian (GEH) which is a special method of clustering the Hessian over subregions of the input manifold.
In this way, they approximate the expectation ~\eqref{eq:grad_interaction} while accounting for error in a high-noise environment.
This is the reason they chose a probabilistic framework with BNNs.
This work is able to get a better handle on the type 1 and type 2 errors made on the interactions when the data is inherently noisy.
Moreover, they are able to show that interaction detection quality directly correlates with the BNN's estimation performance unlike NID which only assumes such a property.
The concern with this method is regarding the ability to train a BNN's estimation to sufficient accuracy given modern variational inference.

\subsection{Feature Interaction Interpretation}
\label{sec:interpretation}

We now overview various approaches to interpreting feature interactions: interpreting interaction coefficients and plots in \S\ref{sec:coefficient} and interaction attribution in \S\ref{sec:attribution}.

\subsubsection{Interpreting Coefficients and Plots}
\label{sec:coefficient}
Various methods exist on how to interpret a given feature interaction. 
One of the simplest approaches is in a multiple regression setting, where a linear regression model is used to capture multiplicative interactions.
For two features, let such a model be defined as
\begin{align}
f(\V{x}) = w_1x + w_2z + w_{3}xz + b,
\label{eq:xz}
\end{align}
where $x$, $z$ are the features, $\{w_i\}$ are their coefficients, and $b$ is a bias term. \citet{jaccard2003interaction} suggest two ways to interpret the $xz$ interaction. One is by directly examining the $w_{3}$ coefficient as a slope varying with $x$ when $z$ increases by one unit. Another way is by rearranging~\eqref{eq:xz} as
\begin{align}
f(\V{x}) = (w_1 + w_{3}z) x + (w_2z + b),
\end{align}
and interpreting $(w_1 + w_{3}z)$ as a slope for fixed values of $z$.~\citet{aiken1991multiple} recommend using this alternative slope to plot interaction effects for representative values of $z$.  The differences in these slopes is also a measure of the significance of the interaction. 

In modern applications, this is most commonly used in conjunction with a local interpretation method like LIME \cite{ribeiro2017lime} to get an idea of how the larger model thinks the two features are interacting.

A more recent work developed an extension to generalized additive model (GAM) to account for feature interactions.
~\citet{lou2013accurate} introduced Generalized Additive Models with Pairwise Interactions (GAM2) to add models of pairwise interactions to the GAM in the form of $g(E[y])=\sum{f_i(x_i)} + \sum{f_{ij}(x_i,x_j)}$.
Each of the $f_i$ and $f_{ij}$ can be of any possible complexity such as a random forest or a neural network.
In theory, a GAM can be extended to interactions of any possible order, which would result in a normal, unrestricted model.
The reason GAM2 stays at only two degree interactions is because the resulting model remains almost fully interpretable.
~\citet{lou2013accurate} and~\citet{caruana2015intelligible} interpreted the interaction effects by plotting each $f_{ij}$ fully as a heatmap.
Higher order interactions do not have such easily visualisable methods which would limit the model's corresponding interpretability.


\subsubsection{Interpreting Interaction Attribution}
\label{sec:attribution}

With the onset of deep learning and feature attribution research, several methods were developed to compute attribution scores for feature interactions. 
These attribution scores apply to \emph{individual data instances} and estimate the impact of interactions on a prediction.
Because this methods focus on interpreting a single data instance, this domain comes with a diverse set of different explanation structures with potentially unaligned goals.

\paragraph{Shapley-Taylor Index:}
The Shapley Taylor Interaction Index (STI)~\citep{dhamdhere2019shapley} extends the Shapley Index, an interpretability method which attempts to `distribute' a decision to each of its features borrowed from coalitional game theory.
This method extends Shapley index from main effects to interaction effects by adding higher order Taylor series terms to the Shapley calculation.
They manage to maintain all of the axiomatic properties of Shapley index when every input features is considered to be a binary feature.
Only working on discrete inputs can be considered a serious weakness, but a continuous extension, while unlikely to satisfy all the original axioms, is likely to exist.
Another issue with Shapley index is that it generally requires all $\mathcal{O}(2^p)$ calculations in order to get the exact Shapley index and there is currently no great understanding of how much sampling should be done to get an accurate estimate of the exact indices.

\paragraph{Integrated Hessians: }
The Integrated Gradients (IG)~\citep{sundararajan2017axiomatic} is an extremely popular path integration method used for feature attribution.
It obeys a host of axioms similar to Shapley index while also being applicable to continuous input variables.
It uses the following equation to integrate over the model's decision output F from a baseline input x' to the input of interest x with a linear interpolation.
\[
IG_i(x) = (x_i -x'_i) \int_{\alpha=0}^1 \frac{\partial }{\partial x_i} F(x' + \alpha (x-x')) \mkern6mu d\alpha
\]
Because it works on any differentiable model, it has become a staple for interpretation in some deep learning communities.
The Integrated Hessians (IH) method~\citep{janizek2020explaining} extends this to interactions by applying the method to itself.
In this way, the method describes how important j's importance is to i, but is more easily understood as the feature interaction.
\[
IH_{i,j}(x) = (x_i -x'_i)(x_j -x'_j) \int_{\beta=0}^1 \int_{\alpha=0}^1 \alpha\beta\cdot \frac{\partial^2 }{\partial x_i\partial x_j} F(x' + \alpha\beta (x-x')) \mkern8mu d\alpha d\beta
\]
This method is able to enjoy a lot of the axioms coming from IG, but one of its major drawbacks is a result of not formalizing interactions and main effects.
The term $IH_{i,i}$ is more like a residual term instead of the main effect, which can make it difficult to identify how much of the actual effects are being properly captured by this method.

\paragraph{Model-Agnostic Hierarchical Explanations:}
Model-Agnostic Hierarchical Explanations (MAHE)~\citep{tsang2018can} trains a surrogate explainer models for interaction detection and attribution in a method similar to LIME.
MAHE's explanations are then fed through a NID model and final GAM to get the attributions.
This methodology allows MAHE to provide explanations relatively quickly by not needing to look at the overall feature space.
Because MAHE is trained on overlapping feature sets, MAHE can fail a lot of the desirable axioms satisfied by these other attribution methods.

\paragraph{Agglomerative Contextual Decomposition:}
Agglomerative Contextual Decomposition (ACD) ~\citep{singh2018hierarchical} is a hierarchical extension of Contextual Decomposition~\citep{murdoch2018beyond} which was originally designed for LSTMs.
This method extended the additive decomposition to DNNs in general by adding technical solutions for the method to work on convolutions and maxpooling, both common operations in CNNs.
The method decomposes each layer as:
\[
g_i(x) = \beta_i(x) + \gamma_i(x)
\]
Each $g_i$ denotes a layer of the network and where $\beta$ denotes the contribution only by the features of interest and $\gamma$ denotes the residual contribution for the model.
Thus, the $\beta$ values can be seen as the interaction effect of a certain group of features when it is the final $\beta$ value from the logit/ prediction layer.
The method then accumulates these scores into a tree hierarchy over the input.
Because this work was designed for text and images, this looks something like a tree of word subsets with their corresponding polarity or image patches with their corresponding classes.
A primary weakness is this method also fails many axiomatic interaction scores because of its formulation.
A minor weakness is the computation time required by the iterative, hierarchical, full-explanation approach.
This is because a partial hierarchy could be computed if desired and many pieces of this framework are flexible for quicker or even different inference.

\paragraph{Archipelago:}
Archipelago~\citep{tsang2020does} also uses a two stage attribution procedure.
First, this method applies ArchDetect which identifies the features which are locally affecting the prediction by using an approximation of the Hessian.
Second, this method applies ArchAttribute which identifies the impact each interaction has on the prediction compared to a pre-defined baseline.
The Hessian detection is done with the following secant approximation:
\[
\omega_{i,j}(\V{x})= \left(\tfrac{1}{h_i h_j} \left(f(\V{x}^{\star}_{\{i,j\}} +  \V{x}_{\setminus{\{i,j\}}}) - f(\V{x}'_{\{i\}} + \V{x}^{\star}_{\{j\}} +  \V{x}_{\setminus{\{i,j\}}}) - f(\V{x}^{\star}_{\{i\}} + \V{x}'_{\{j\}} +  \V{x}_{\setminus{\{i,j\}}}) + f(\V{x}'_{\{i,j\}} +   \V{x}_{\setminus{\{i,j\}}})\right)\right)
\]
\[
\text{where}\quad
(\V{x}_{\II})_i = \left\{\begin{array}{lr}
        x_i, & \text{if } i\in \II\\
        0 & \text{otherwise }\\
        \end{array}\right.
\]
This approximant is then accumulated over samples in the dataset within multiple contexts $x^*$ which leads to the final detections.
These detected interactions are given their final attribution which is done with the following equation:
\[
\phi(\II) = f(\V{x}^{\star}_\II+\V{x}'_{\setminus \II}) - f(\V{x}').
\]
Because of Archipelago's simple but effective formulation, it is able to perform inference quicker than all the previously mentioned methods while still being able to satisfy the desired interpretability axioms of the aforementioned methods.

\paragraph{Overview of Axiom Satisfaction:}
The following table helps give a brief overview of which axioms are satisfied by which algorithms.
\begin{table}[htb]
      \centering
     \resizebox{1\columnwidth}{!}{%
    \begin{tabular}{lcccccc}
    Methods & Completeness & Set Attribution & Sensitivity & Implementation Invariance & Linearity & Symmetry-Preserving\\
    SCD$\ddagger$~\cite{jin2019towards} &$\times$&$\times$&$\times$&$\times$&$\times$&$\times$\\
    CD$\ddagger$~\cite{murdoch2018beyond,singh2018hierarchical}&$\times$&$\times$&$\times$&$\times$&$\times$&$\times$\\
    SOC~\cite{jin2019towards}&$\times$&$\times$&$\times$&\checkmark&$\times$&$\times$\\
    SI$\ddagger$~\cite{grabisch1999axiomatic} &$\times$&$\times$&$\times$&\checkmark&$\times$&$\times$\\
    IG$\dagger$~\cite{sundararajan2017axiomatic} &\checkmark&$\times$&\checkmark&\checkmark&\checkmark&\checkmark\\
    SHAP~\cite{lundberg2017unified} &\checkmark&$\times$&$\times$&\checkmark&$\times$&$\times$\\
    IH~\cite{janizek2020explaining}&\checkmark&$\times$&\checkmark&\checkmark&\checkmark&\checkmark\\
    STI~\cite{dhamdhere2019shapley} &\checkmark&$\times$&$\times$&\checkmark&$\times$&$\times$\\
    Archipelago~\cite{tsang2020does}&\checkmark&\checkmark&\checkmark&\checkmark&\checkmark&\checkmark\\
    \multicolumn{7}{l}{\footnotesize $\dagger$ IG~\cite{sundararajan2017axiomatic} under specific conditions can be enabled to follow the Set-Attribution axiom.}\\
    \multicolumn{7}{l}{\footnotesize $\ddagger$ The authors do not show the applicability of axioms.}\\
 \end{tabular}%
    }
    \caption{Methods with Axiom Satisfaction}
    \label{table:relatedworks}

\end{table}

\section{Applications}
\label{sec:application}

We overview several applications relevant to feature interactions for interpretable machine learning.
Since deep learning has become mainstream, there have been efforts on interpreting or leveraging feature interactions that are captured by deep neural networks. 
Here, we focus our discussions on research related to three types of models:  text analyzers, image classifiers, and recommender systems. Since a feature interaction is a general phenomenon, we also discuss applications of high interest to scientific communities.

\subsection{Text Analyzers}

A prominent interest of the text analysis community is explaining word interactions in applications like sentiment analysis. The explanations can indicate how words modify each others’ sentiment when considered together rather than separately. On this topic,~\citet{murdoch2018beyond} proposed Contextual Decomposition (CD) to extract word interactions from Long Short-Term Memory (LSTM) networks~\citep{hochreiter1997long}  in the form of word-phrase attributions decomposed throughout the network.~\citet{jin2019towards} advanced CD with a method called Sampling Contextual Decomposition (SCD), which applies word sampling to the neural activation decomposition of CD. In addition,~\citet{jin2019towards}  proposed Sampling Occlusion (SOC), which samples words around a target phrase in occlusion-based attribution scoring. SCD and SOC have been applied to both LSTM and state-of-the-art BERT~\citep{devlin2019bert} models. While CD, SCD, and SOC compute attribution scores, they do not detect feature interactions nor apply to general sets of words.

Shapley Taylor Interaction Index~\citep{dhamdhere2019shapley}, Integrated Hessians~\citep{janizek2020explaining}, MADEX~\citep{tsang2020feature}, and Archipelago~\citep{tsang2020does} were interaction-based methods  applied to text analysis, with the latter method addressing attribution interpretability issues as mentioned earlier.

\subsection{Image Classifiers}

Limited works have studied how to interpret feature interactions in image classifiers. In~\citet{singh2018hierarchical}, the Contextual Decomposition (CD) method was expanded to image classification, but CD still does not detect interactions nor are its attributions axiomatic in the sense of~\citet{sundararajan2017axiomatic}.  Some methods attempt to interpret feature groups in image classifiers, such as Anchors~\citep{ribeiro2018anchors}, and Context-Aware methods~\citep{singla2019understanding}; however, these methods face the same drawbacks as CD. MADEX~\citep{tsang2020feature} is a black-box interaction detection method applied to image classifiers, whereas Archipelago~\citep{tsang2020does} was a follow-up method that also addressed interaction attribution in images. 

A significant issue for interaction interpretation of image classifiers is that images are extremely high-dimensional inputs, making pixel interaction detection very slow and impractical. Existing methods circumvented this problem by only examining image segment (superpixel) interactions rather than pixel interactions~\citep{tsang2020feature,tsang2020does}.

\subsection{Recommender Systems}

The goal of recommender systems is to provide personalized recommendations to users by leveraging user-item interactions. Here, ``item'' refers to the item being recommended. Despite the importance of user-item interactions, limited works studied how to interpret interactions in recommender systems,  instead focusing on modeling interactions for prediction. For example,~\citet{cheng2016wide},~\citet{guo2017deepfm},~\citet{wang2017deep}, and~\citet{lian2018xdeepfm} directly incorporate multiplicative cross terms in neural network architectures, and~\citet{song2018autoint} use attention as an interaction module, all of which are intended to improve the neural network's function approximation. 
The attention module in~\citet{song2018autoint} was motivated in part for interpretability, but it remains to been seen how interpretable attention modules are.
GLIDER~\citep{tsang2020feature} proposed global feature interaction detection and associated modeling in black-box recommender systems, whereas Archipelago~\citep{tsang2020does} applied interaction attribution to these systems.

\subsection{Other Science Applications}
The study of \textbf{genetics} requires a focus on the epistatic interactions between different genes.  
\citet{greenside2018dnaDFIM} uses explicit interactions to bring interpretability to deep models in DNA sequencing.
There is also the physical interaction of a protein binding to a DNA molecule for gene expression.
Many health science applications, particularly \textbf{medicine}, need to understand the interplay between a potential treatment and the patient's diet, behavior, other current drugs, and even genetics to get a full understanding of the risks and the benefits of that treatment.
Graph Networks have made great strides in \textbf{chemistry} by functioning on the individual molecule and treating it as a graph.
For example, the work in ~\citep{sanch2020evalGNN} attempts to bring interpretability to these GNN frameworks which are able to identify complex interactions among atoms within a molecule. 

Practically all sciences have some interactions between entities:
\begin{itemize}
    \item \textbf{Biological Interaction (Ecology):} The effect that a pair of organisms living together in a community have on each other. 
    \item \textbf{Tectonic–Climatic Interaction (Geology):} The  effect of both tectonics and climate patterns on orogenesis, volcanism, and erosion events as well as  atmospheric circulation, orographic lift, monsoon circulation and rain shadow.
    \item \textbf{Physics:} Fundamental interactions between particles like gravity, electromagnetism, strong force, and weak force.
    \item \textbf{Neuron Interaction (Neuroscience)}: The effect of multiple neurons on memory, learning, and reasoning and the way in which they culminate to produce these effects.
\end{itemize}

It is possible to study almost all of these interactions with the tools discussed in this survey. 
Given the existing tools, the challenge of studying these scientific applications is often in data collection. 
However, interaction analysis on spatio-temporal models for these applications may present interesting interpretable machine learning problems.

\section{Opportunities}
\label{sec:future}

For future research, there are many possible directions to go with this diversely applicable field, but we also offer the following list of recommendations. 
We suggest new research into alternative ways of explaining feature interactions. Despite the importance of interactions involving multiple feautres, they are often difficult to visualize and understand.
Hence there is a need to communicate such interactions via advances in visualization or mathematics. 
In addition, we suggest new investigation into the intersection of interaction explanations and causal inference. 
This intersection presents the question of how does a treatment variable interact with existing features for the prediction of a causal effect. 
The features involved in the interaction could reveal how treatment effects personalize to subgroups of users by leveraging explanations of modern prediction models.

User experience researchers may also be interested in  studying ways to best convey feature interactions, such as through interactive visualizations.
The insights drawn from the feature interaction analysis can be useful to researchers in practically any field where feature interaction discovery has fundamental importance.
In order to better validate the interpretations by new or existing interaction explanation methods, we advocate the development of automatic evaluation metrics and datasets as well as advanced user studies for qualitative evaluations.

\bibliography{refs}
\bibliographystyle{abbrvnat}

\end{document}